# Second Order Swarm Intelligence

Vitorino Ramos[1], David M.S. Rodrigues[2,3], and Jorge Louçã[3]

[1]*LaSEEB* – Evolutionary Systems and Biomedical Eng. Lab., ISR – Robotic and Systems Institute, Technical University of Lisbon (IST), Av. Rovisco País, 1 Torre Norte, 6.21, 1049-001 Lisbon, PORTUGAL
vitorino.ramos@ist.utl.pt
[2]The Open University, Milton Keynes, UNITED KINGDOM
david.rodrigues@open.ac.uk
[3]*The Observatorium* - ISCTE-IUL, Lisbon University Institute (IUL), Av. Forças Armadas, 1649-026 Lisbon, PORTUGAL
jorge.l@iscte.pt

**Abstract**: An artificial Ant Colony System (ACS) algorithm to solve general-purpose combinatorial Optimization Problems (COP) that extends previous AC models [21] by the inclusion of a negative pheromone, is here described. Several Traveling Salesman Problem (TSP) were used as benchmark. We show that by using two different sets of pheromones, a second-order coevolved compromise between positive and negative feedbacks achieves better results than single positive feedback systems. The algorithm was tested against known NP-complete combinatorial Optimization Problems, running on symmetrical TSPs. We show that the new algorithm compares favorably against these benchmarks, accordingly to recent biological findings by Robinson [26,27], and Grüter [28] where "No entry" signals and negative feedback allows a colony to quickly re-allocate the majority of its foragers to superior food patches. This is the first time an extended ACS algorithm is implemented with these successful characteristics.

## 1 Introduction

Research over hard NP-complete *Combinatorial Optimization Problems* (COP's) has, in recent years, been focused on several robust bio-inspired meta-heuristics, like those involving *Evolutionary Computation* (EC) algorithmic paradigms [1-3], as well as other kind of heuristics and approximation algorithms [4-5]. One particularly successful well-know meta-heuristic [6] approach is based on *Swarm Intelligence* (SI) [7-8], i.e., the self-organized stigmergic-based [9-11] property of a complex system whereby the collective behaviors of (unsophisticated) entities interacting locally with their en-vironment cause coherent functional global patterns to emerge [12]. This line of research recognized as *Ant Colony Optimization* (ACO) [13-15], uses a set of stochastic cooperating ant-like agents to find good solutions, using self-organized *Stigmergy* [16-19] as an indirect form of communication mediated by an artificial pheromone, whereas agents deposit pheromone-signs on the edges of the problem-related complex network, encompassing a family of successful algorithmic variations such as: *Ant Systems* (AS) [20], *Ant Colony Systems* (ACS) [21], *Max-Min Ant Systems* (Max-Min AS) [22] and Ant-*Q* [23].

Albeit being extremely successful these algorithms mostly rely on positive feedbacks [13], causing excessive algorithmic exploitation over the entire combinatorial search space. This is particularly evident over well-known benchmarks as the symmetrical *Traveling Salesman Problem* (TSP) [24]. Being these systems comprised of a large number of frequently similar components or events, the main challenge is to understand how the components interact to produce a complex pattern that is still a feasible solution [25] (in our case study, an optimal robust solution for hard NP-complete dynamic TSP-like combinatorial problems).

In order to overcome this hard search space exploitation-exploration compromise, our present algorithmic approach follows the route of very recent biological findings [26-28] showing that forager ants lay attractive trail pheromones to guide nest mates to food, but where, the effectiveness of foraging networks were improved if pheromones could also be used to repel foragers from unrewarding routes. Increasing empirical evidences for such a negative trail pheromone exists, deployed by *Pharaoh*'s ants (*Monomorium pharaonis*) as a '*no entry*' signal to mark unrewarding foraging paths.

The new algorithm was exhaustively tested on a series of well-known benchmarks over hard NP-complete COP's, running on symmetrical TSP [24]. Different network topologies and stress tests were conducted over low-size TSP's, medium-size TSP's, and as well as large sized ones. We show that the new co-evolved stigmergic algorithm compared favorably against the benchmark. In order to deeply understand how a second co-evolved pheromone was useful to drive the collective system into such results, the influence of negative pheromone was tracked (fig. 3-4-5), and as in previous tests [29-30], a refined phase-space map was produced mapping the pheromones ratio between a pure Ant Colony System and the present second-order approach.

## 2 Towards a Co-Evolving Swarm-Intelligence

In order to make use of co-evolution we created a double-pheromone model on top of the traditional ACS, thus allowing the comparison between the two, by having an additional parameter. Traditional approaches to the TSP via Ant Systems include only a positive reinforcement pheromone. Our approach uses a second negative pheromone, which acts as a marker for forbidden paths. These paths are obtained from the worse tour of the ants and this pheromone then blocks access of ants in subsequent tours. This blockade isn't permanent and as the pheromone evaporates it allows paths to be searched again for better solutions. This leads to equations 5-9 that expand equations 1-4 of the original ACS and AS approaches.

***Ant Colony System* (ACS, [21]) *state transition* rule**

$$s = \begin{cases} \operatorname{argmax}_{u \in J_k(r)} \left\{ [\tau(r,u)] \cdot [\eta(r,u)]^{\beta} \right\}, & \text{if } q \le q_0 \text{ (exploitation)} \\ S, & \text{otherwise (biased exploration} \end{cases} \tag{1}$$

***Ant System* (AS, [20]) *random proportional* rule**

$$p_k = \begin{cases} \dfrac{[\tau(r,s)] \cdot [\eta(r,s)]^{\beta}}{\sum_{u \in J_k(r)} [\tau(r,u)] \cdot [\eta(r,u)]^{\beta}}, & \text{if } s \in J_k(r) \\ 0, & \text{otherwise} \end{cases} \tag{2}$$

***Ant Colony System* (ACS, [21]) *local updating* rule**

$$\tau(r,s) \leftarrow (1-\rho) \cdot \tau(r,s) + \rho \cdot \Delta\tau(r,s) \tag{3}$$

***Ant Colony System* (ACS, [21]) *global updating* rule**

$$\tau(r,s) \leftarrow (1-\alpha) \cdot \tau(r,s) + \alpha \cdot \Delta\tau(r,s) \tag{4}$$

### 2.1 ACS double-pheromone state transition rule

Following the guidelines of Dorigo and Gambardella [21], in ACS the state transition rule is as follows: an ant positioned on node *r* chooses the city *s* to move to by applying the rule given in Eq.5

$$s = \begin{cases} \operatorname{argmax}_{u \in J_k(r)} \left\{ [\tau^{+}(r,s)]^{\alpha} \cdot [\eta(r,s)]^{\beta} \cdot [\tau^{-}(r,s)]^{\alpha-1} \right\}, & \text{if } q \le q_0 \\ S, & \text{otherwise} \end{cases} \tag{5}$$

where $q$ is a random number uniformly distributed in [0...1], $q_0$ is a parameter ($0<=q_0<=1$) and $S$ is a random variable selected according to the probability distribution in Eq. 6. In Eq.5, the parameter $q_0$ determines the relative importance of exploitation versus exploration, that is, every time an ant in city $r$ has to choose a city $s$ to move to, it samples a random number between $0<=q_0<=1$. If $q<=q_0$ then the best edge according to Eq.5 is chosen (exploitation), otherwise an edge is chosen according to Eq.6 (biased exploration) or *random-proportional rule* coming from the classic *Ant System* (AS), which follows:

$$p_k = \begin{cases} \dfrac{\left[\tau^+(r,s)\right]^{\alpha} \cdot \left[\eta(r,s)\right]^{\beta} \cdot \left[\tau^-(r,s)\right]^{\alpha-1}}{\sum_{u \in J_k(r)} \left[\tau^+(r,u)\right]^{\alpha} \cdot \left[\eta(r,u)\right]^{\beta} \cdot \left[\tau^-(r,u)\right]^{\alpha-1}}, & \text{if } s \in J_k(r) \\ 0, & \text{otherwise} \end{cases} \qquad (6)$$

Eq.6 gives the probability with which ant $k$ in city $r$ chooses to move to city $s$, where $\tau$ is the pheromone on the $(r,s)$ network edge, $\eta=1/\delta$ is the inverse of the distance $\delta(r,s)$, $J_k(r)$ is the set of cities that remain to be visited by ant $k$ positioned on city $r$ (in order to make the solution feasible), and $\beta$ is a parameter which determines the relative importance of pheromone versus distance ($\beta>0$) and $\alpha$ controls the ratio between positive and negative pheromone influences. In Eq.6 the pheromones on edge $(r,s)$ are multiplied by the corresponding heuristic value $\eta(r,s)$, thus favoring the choice of edges which not only are shorter but also with a greater amount of positive pheromone and some amount of negative pheromone.

The final ACS state transition rule resulting from Eqs. 5 and 6 is then called *pseudo-random-proportional rule*. This state transition rule, as with the previous AS random-proportional rule, favors transitions towards nodes connected by short edges and with a large amount of pheromone.

### 2.2 ACS double-pheromone global updating rule

While AS used $L_k$, the length of the tour performed by every ant $k$, as a heuristic measure for the pheromone global updating rule, ACS instead focus only in the globally best ant, among all $m$, i.e. the ant which constructed the shortest tour from the beginning of the trial is allowed to deposit pheromone. This choice, along with the use of the *pseudo-random-proportional* state transition rule (above) was intended to make the search more directed: ants search in a neighborhood of the best tour found up to the current iteration of the algorithm. Global updating is performed after all ants have completed their tours. The pheromone level is then updated by applying the global updating rule of Eq.7 and 8 below:

$$\tau^+(r,s) \leftarrow (1-\rho^+) \cdot \tau^+(r,s) + \rho^+ \cdot \Delta\tau^+(r,s) \qquad (7)$$

$$\Delta\tau^+(r,s) = \begin{cases} (L_{gb})^{-1}, & \text{if } (r,s) \in \text{ Global best tour} \\ 0, & \text{otherwise} \end{cases}$$

where $0< \rho^{\pm} <1$ is the pheromone decay parameter (evaporation) and $L_{gb}$ the length of the globally best tour from the beginning of the trial. As it was the case in AS, the ACS global pheromone updating provides a greater amount of pheromone to shorter tours. Eq.7 dictates that only those edges belonging to the global best tour will receive reinforcement while Eq.8 dictates that only those edges that belong to the worse tour receive negative pheromone deposition.

### 2.3 ACS double-pheromone local updating rule

In order for the 2nd order algorithm (as in ACS) to build a solution, i.e. a TSP tour, ants visit edges and change their pheromone level by applying a local updating rule given by Eq.9:

$$\tau^{-}(r,s) \leftarrow (1-\rho^{-})\cdot\tau^{-}(r,s)+\rho^{-}\cdot\Delta\tau^{-}(r,s) \tag{8}$$

$$\Delta\tau^{-}(r,s)=\begin{cases} (nL_{gb})^{-1}, \text{ if } (r,s)\in \text{ Global worse tour} \\ 0, \qquad \text{otherwise} \end{cases}$$

$$\begin{aligned} \tau^{+}(r,s) &\leftarrow (1-\rho^{+})\cdot\tau^{+}(r,s)+\rho^{+}\cdot\Delta\tau^{+}(r,s) \\ \tau^{-}(r,s) &\leftarrow (1-\rho^{-})\cdot\tau^{-}(r,s)+\rho^{-}\cdot\Delta\tau^{-}(r,s) \end{aligned} \tag{9}$$

where $0<\rho<1$ is a parameter. From here several options are possible where, $\Delta\tau(r,s)$ could assume the form of $\Delta\tau(r,s)=\gamma$. max $\tau(s,z)$ [$z$ . $J_k(S)$ ] similarly to a reinforcement learning problem, onto which ants have to learn which city to move to as a function of their current location. This first option assumes $Q$-learning, an algorithm which allows an agent to learn such an optimal policy by the recursive application of a rule similar to that in Eq.4, giving rise to the first Ant-$Q$ ant systems. In fact, $\Delta\tau(r,s)=\gamma$. max $\tau(s,z)$ is exactly the same formula used in $Q$-learning where $0<\gamma<1$ is a parameter. The other two choices are normally: (1) setting $\Delta\tau(r,s)=\tau_0$, being $\tau_0$ the initial pheromone level, or (2) simply setting $\Delta\tau(r,s)=0$. Finally, experiments could also be ran in which local-updating are not applied at all, that is, where the local updating rule is not used as in the case of the older and previous AS).

Current research work however, suggests that local-updating is not only definitely useful, but that the pheromone local updating rule with $\Delta\tau(r,s)=0$ yields worse performance than $\Delta\tau(r,s)=\tau_0$ or even Ant-$Q$. In fact, $\Delta\tau(r,s)=\tau_0$ was chosen for the standard ACS, from the beginning [13-15][20,21,23]. Since the ACS local updating rules not only requires less computation than Ant-$Q$ as well as achieving better results, we chose to focus our attention on ACS, which will be used, along others, to run the comparison experiments against our new co-evolved pheromone-based algorithm in the following paper sections.

## 3 Results

The new algorithm was exhaustively tested on a series of well-known benchmarks over hard NP-complete COP's) running on symmetrical TSP's. Different network topologies and stress tests were conducted over low-size TSP's (eil51.tsp; eil78.tsp; kroA100.tsp), medium-size (d198.tsp; lin318.tsp; pcb442.tsp; rat783.tsp) as well as large sized ones (fl1577.tsp; d2103.tsp) [numbers here referring to the number of nodes in each network].

**Table 1.** Comparison of Standard ACS with the 2nd order AS algorithm

| *problem* | *n.º of nodes* | *Standard ACS* | *2nd order+ AS* | *optimal tour* |
|---|---|---|---|---|
| eil51.tsp | 51 | 427.96 | 428.87 | 426 |
| eil78.tsp | 78 | ** | 544.65 | 538 |
| kroA100.tsp | 100 | 21285.44 | **21285.44** | 21282 |
| d198.tsp | 198 | 16054 | **15900.2** | 15780 |
| lin318.tsp | 318 | 42029*** | 42683.90 | 42029 |
| pcb442.tsp | 442 | 51690 | **51464.48** | 50778 |
| rat783.tsp | 783 | 9066 | **8910.48** | 8806 |
| fl1577.tsp | 1577 | 23163 | **22518** | 22249 |
| d2103.tsp | 2103 | - | 81151.9 | 80450 |

All optimal tours from http://comopt.ifi.uni-heidelberg.de/software/TSPLIB95/STSP.html

+ Average over 20 runs and limited to 1000 iterations

** Value for similar problem eil75,.tsp - 542.37 *** uses 3-opt local search

Comparing Traditional AS Models with 2nd Order

It is clear from table 1 that the 2nd order AS performs at least equally, if not better, than the standard ACS. It is clearly seen the averages of the runs (bold) that are better than the traditional ACS.

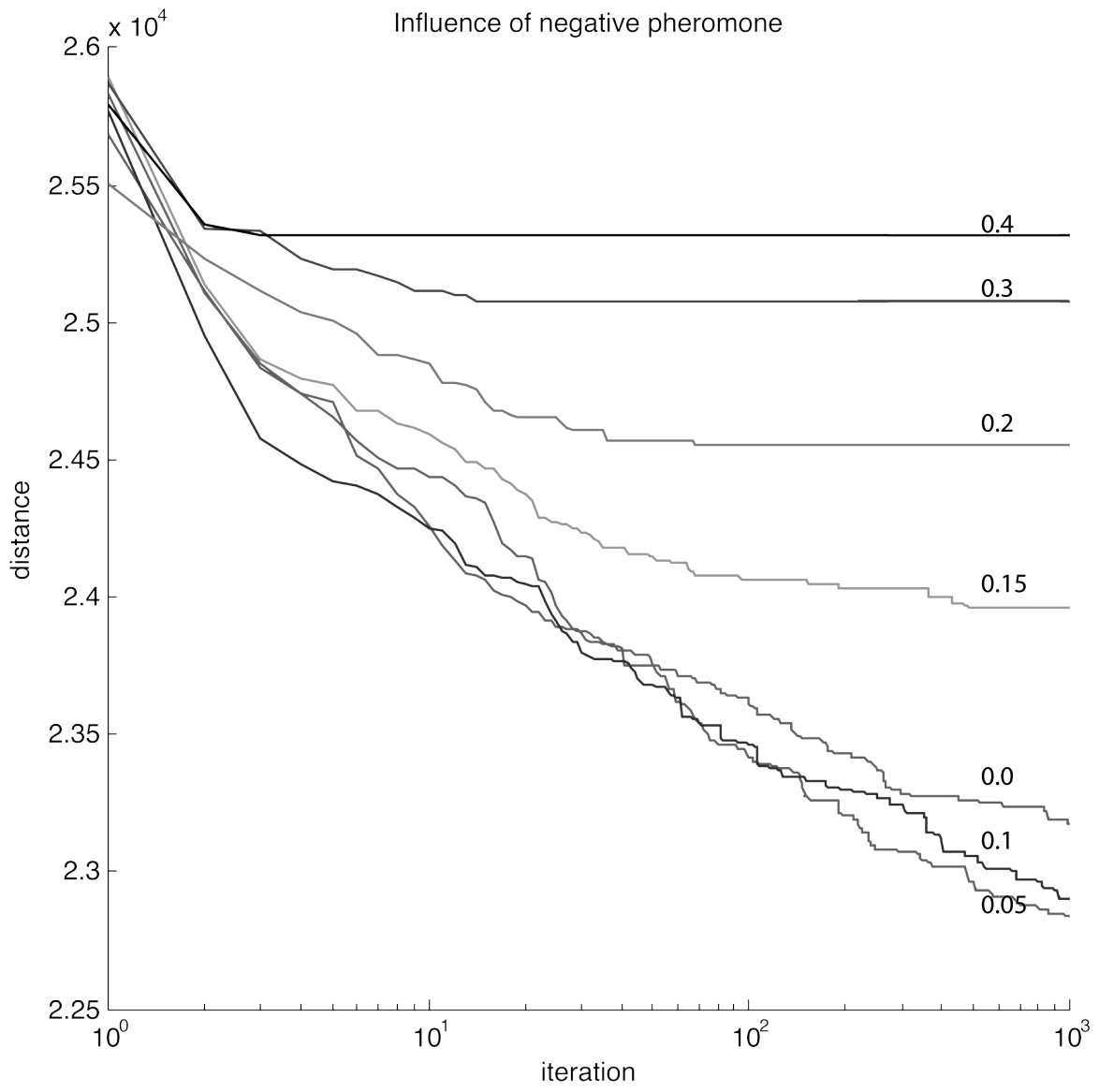

**Fig. 1.** Influence of negative pheromone on *kroA100.tsp* problem (values on lines represent 1-ALPHA)

We investigated the evolution of different ratios of negative pheromone and found that a small amount of negative pheromone applied as a non-entry signal indeed produces better results, but the effect is cancelled if the ratio of the negative pheromone is high when compared to the positive pheromone.

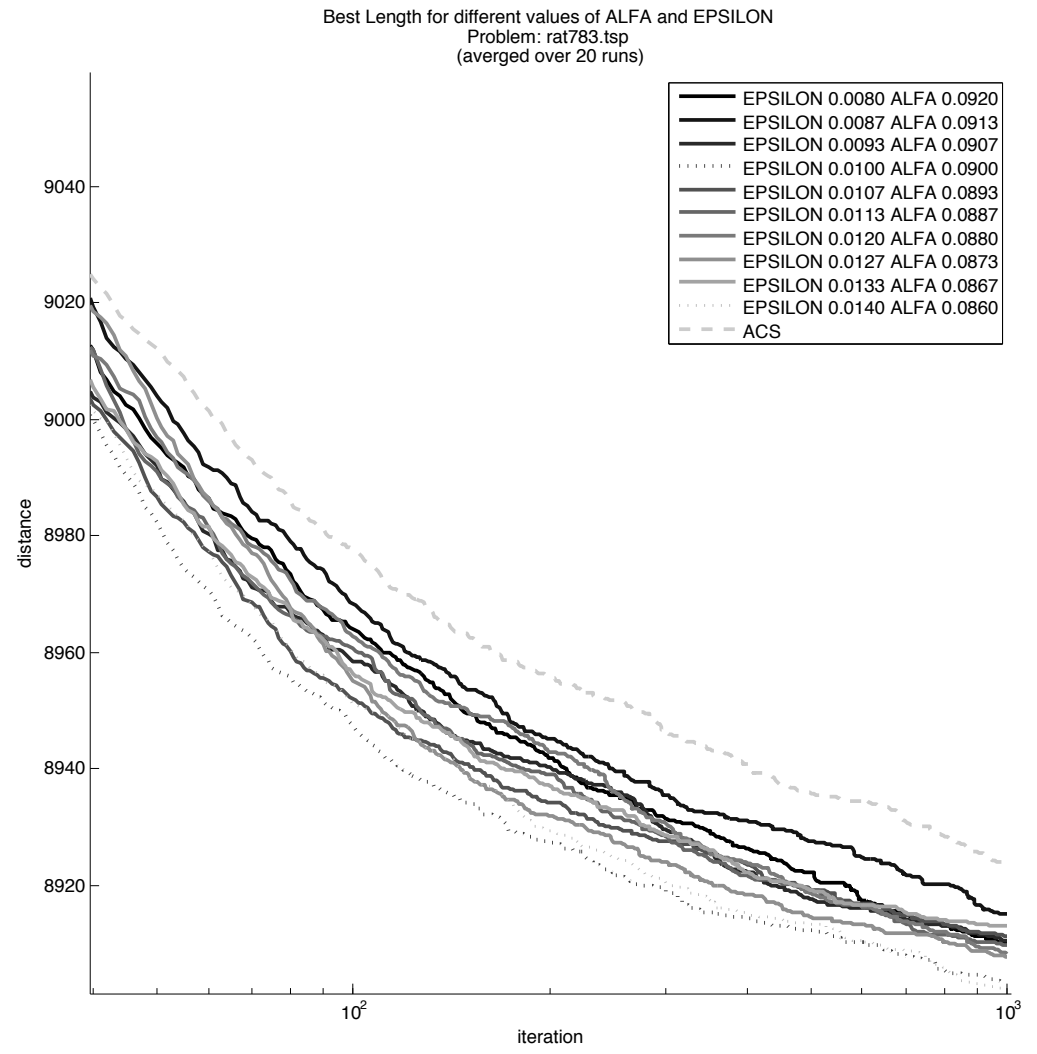

**Fig. 2.** Best tour of the 2nd order AS for different ratios of negative pheromone in the *rat783.tsp* problem

The effect of negative pheromone can be observed both in figure 1 and figure 2 where one can observe that small amounts of negative pheromone produce better results and quicker convergence to those results. On the other hand if one increases the ratio of negative pheromone to higher values then it isn't possible to ripe the benefits of the no-entry signal and the system performs worse.

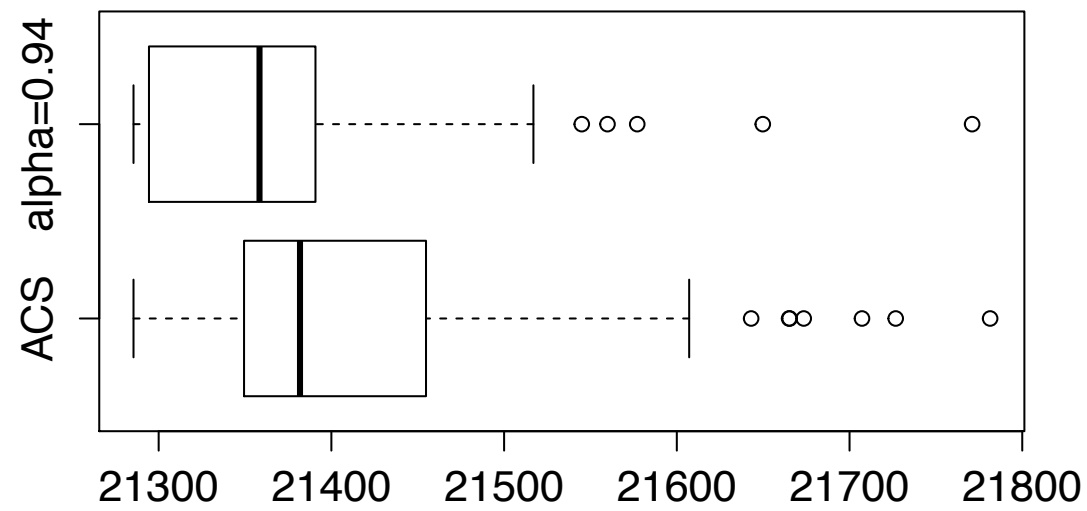

**Fig. 3.** Influence of negative pheromone on *kroA100.tsp* problem.

The detailed analysis of the *kroA100.tsp* problem showed that this effect is statistically significant. Comparing 120 runs with *alpha*=1 (equivalent to traditional ACS) and *alpha=0.94*, we obtained a *p*-value of $3 \times 10^{-4}$. This result is summarized in fig-ure 3, where one compares traditional ACS with our 2nd order approach.

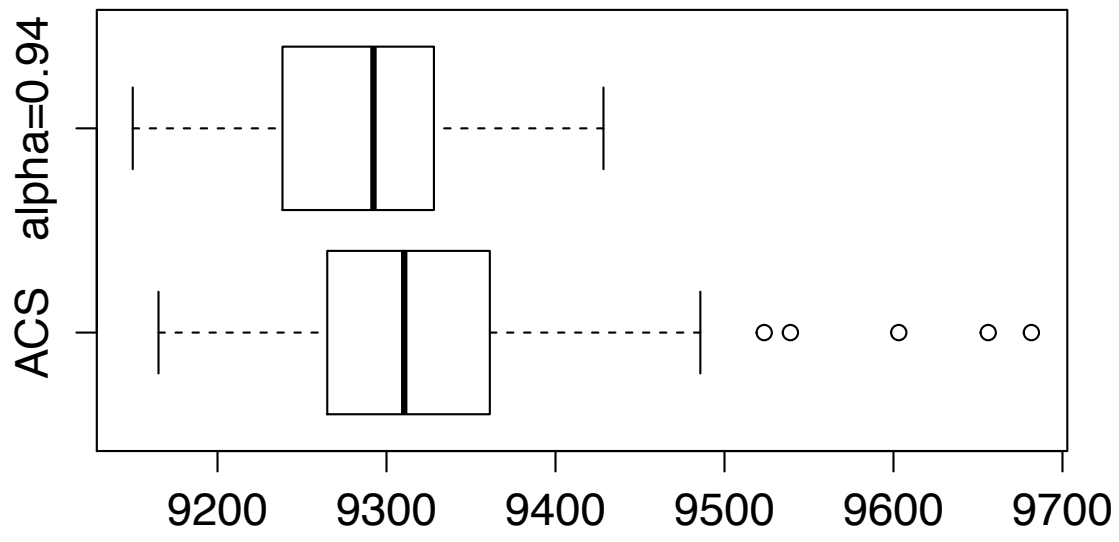

**Fig. 4.** Influence of negative pheromone on *rat783.tsp* problem.

The same results where observed for problem *rat783.tsp* when comparing 70 runs of the ACS (*alpha=1*) with 70 runs of the 2nd order approach (with *alpha=0.94*) in figure 4. The two samples means were tested for statistical significance resulting in a *p*-value of $2.2x10^{-3}$.

Both these examples show that on average the 2nd order approach performs better than traditional ACS. This effect of the negative pheromone is important but cannot be extended further as to dominate the solving strategy, making results worse. This can be seen clearly on figure 5 where further diminishing of *alpha* (giving more weight to negative pheromone as a consequence) produces worse results.

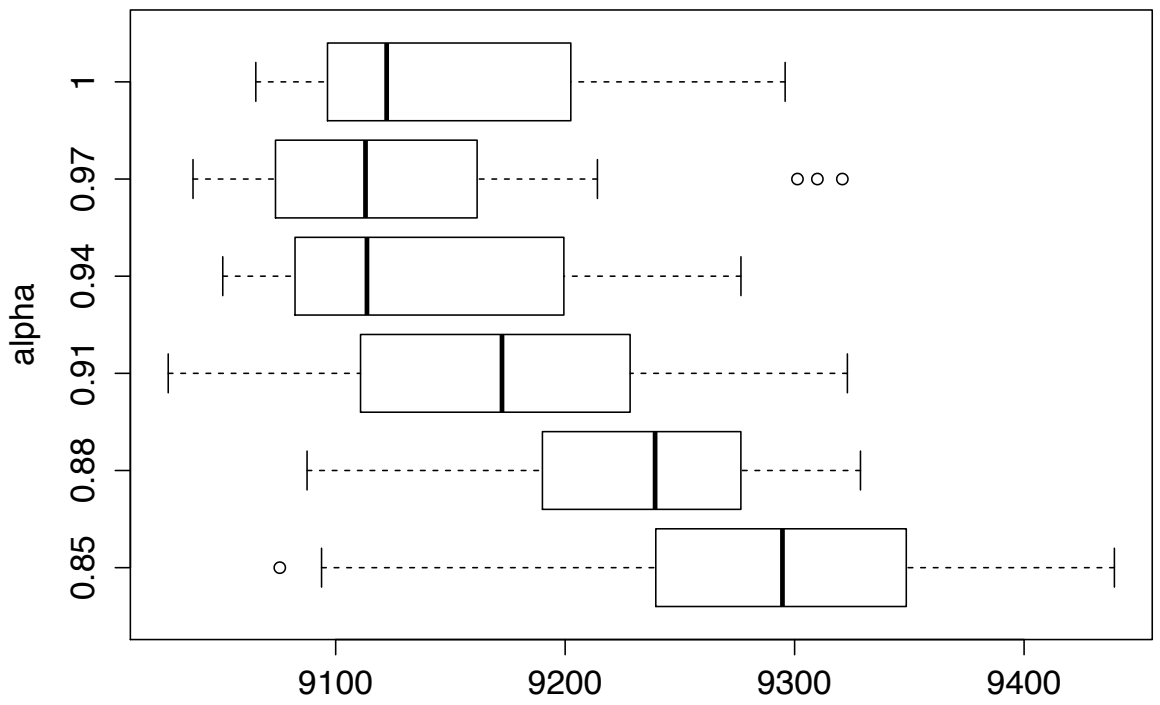

**Fig. 5.** If dominant, negative pheromone has negative impact (problem *rat783.tsp*)

# 4 Conclusion

We show that the new co-evolved stigmergic algorithm compared favorably against the benchmark. The inclusion of a negative pheromone acting as a `non-entry' signal in the strategy of construction of solutions is beneficial as the convergence to optimal solutions is quicker, as shown in figure 2, while achieving better results (figures 3 and 4). The algorithm was able to equal or majorly improve every instance of those standard algorithms.

The new algorithm comprises a second order approach to *Swarm Intelligence*, as pheromone-based no entry-signals cues, were introduced, coevolving with the standard pheromone distributions (collective cognitive maps [12]) in the aforementioned known algorithms.

The use of the negative pheromone is limited to small quantities (*alpha* close to 1, but not 1, in which case we would end up with a pure ACS) and cannot be extended to a point of dominance in the search strategy as shown in figure 5. The results found for the TSP problems in that case are severely worse. This implies that the use of a negative pheromone strategy has to be fine tuned as not to dominate the search strategy. This is done with the introduction of the parameter *alpha* that balances the weight of the two pheromones deposition in equations 5 and 6.

This work has implications in the way large combinatorial problems are addressed as the double feedback mechanism shows improvements over the single-positive feedback mechanisms in terms of convergence speed and of major results.